\documentclass[11pt,a4paper]{article}
\usepackage{hyperref}
\usepackage{xcolor}
\usepackage[nohyperref]{naaclhlt2018}
\usepackage{times}
\usepackage{latexsym}
\usepackage{graphicx}

\usepackage{amsmath}
\usepackage{hhline}

\aclfinalcopy
\renewcommand{\vec}[1]{\mathbf{#1}}

\title{Simple and Effective Multi-Paragraph Reading Comprehension}

\author{
Christopher Clark\Thanks{Work completed while interning at the Allen Institute for Artificial Intelligence} \\
University of Washington \\
\texttt{csquared@cs.washington.edu}
\And 
Matt Gardner \\
Allen Institute for Artificial Intelligence \\
\texttt{mattg@allenai.org}
}

\begin{document}

\maketitle

\begin{abstract}
We consider the problem of adapting neural paragraph-level question answering models to the case where entire documents are given as input.
Our proposed solution trains models to produce well calibrated confidence scores for their results on individual paragraphs.
We sample multiple paragraphs from the documents during training, and use a shared-normalization training objective that encourages the model to produce globally correct output. 
We combine this method with a state-of-the-art pipeline for training models on document QA data.
Experiments demonstrate strong performance on several document QA datasets. Overall, we are able to achieve a score of 71.3 F1 on the web portion of TriviaQA, a large improvement from the 56.7 F1 of the previous best system.
\end{abstract}

\section{Introduction}
Teaching machines to answer arbitrary user-generated questions is a long-term goal of natural language processing. For a wide range of questions, existing information retrieval methods are capable of locating documents that are likely to contain the answer. However, automatically extracting the answer from those texts remains an open challenge. The recent success of neural models at answering questions given a related paragraph~\cite{wang2017gated, tan2017s} suggests neural models have the potential to be a key part of a solution to this problem. Training and testing neural models that take entire documents as input is extremely computationally expensive, so typically this requires adapting a paragraph-level model to process document-level input.

There are two basic approaches to this task. Pipelined approaches select a single paragraph from the input documents, which is then passed to the paragraph model to extract an answer~\cite{triviaqa,wang2017r}. Confidence based methods apply the model to multiple paragraphs and returns the answer with the highest confidence~\cite{openqa}. Confidence methods have the advantage of being robust to errors in the (usually less sophisticated) paragraph selection step, however they require a model that can produce accurate confidence scores for each paragraph. As we shall show, naively trained models often struggle to meet this requirement.

In this paper we start by proposing an improved pipelined method which achieves state-of-the-art results. Then we introduce a method for training models to produce accurate per-paragraph confidence scores, and we show how combining this method with multiple paragraph selection further increases performance.

Our pipelined method focuses on addressing the challenges that come with training on document-level data. We propose a TF-IDF heuristic to select which paragraphs to train and test on. Since annotating entire documents is very expensive, data of this sort is typically distantly supervised, meaning only the answer text, not the answer spans, are known. To handle the noise this creates, we use a summed objective function that marginalizes the model's output over all locations the answer text occurs. 
We apply this approach with a model design that integrates some recent ideas in reading comprehension models, including self-attention~\cite{cheng2016long} and bi-directional attention~\cite{bidaf}.

Our confidence method extends this approach to better handle the multi-paragraph setting. Previous approaches trained the model on questions paired with paragraphs that are known \textit{a priori} to contain the answer. This has several downsides: the model is not trained to produce low confidence scores for paragraphs that do not contain an answer, and the training objective does not require confidence scores to be comparable between paragraphs. We resolve these problems by sampling paragraphs from the context documents, including paragraphs that do not contain an answer, to train on. We then use a shared-normalization objective where paragraphs are processed independently, but the probability of an answer candidate is marginalized over all paragraphs sampled from the same document. This requires the model to produce globally correct output even though each paragraph is processed independently.

We evaluate our work on TriviaQA web~\cite{triviaqa}, a dataset of questions paired with web documents that contain the answer. We achieve 71.3 F1 on the test set, a 15 point absolute gain over prior work. We additionally perform an ablation study on our pipelined method, and we show the effectiveness of our multi-paragraph methods on TriviaQA unfiltered and a modified version of SQuAD~\cite{squad} where only the correct document, not the correct paragraph, is known. We also build a demonstration of our method by combining our model with a re-implementation of the retrieval mechanism used in TriviaQA to build a prototype end-to-end general question answering system~\footnote{documentqa.allenai.org}. We release our code~\footnote{github.com/allenai/document-qa} to facilitate future work in this field.

\section{Pipelined Method}
\label{sect:paragraph-level-models}
In this section we propose an approach to training pipelined question answering systems, where a single paragraph is heuristically extracted from the context document(s) and passed to a paragraph-level QA model. We suggest using a TF-IDF based paragraph selection method and argue that a summed objective function should be used to handle noisy supervision. We also propose a refined model that incorporates some recent modeling ideas for reading comprehension systems. 

\subsection{Paragraph Selection}
Our paragraph selection method chooses the paragraph that has the smallest TF-IDF cosine distance with the question. Document frequencies are computed using just the paragraphs within the relevant documents, not the entire corpus. The advantage of this approach is that if a question word is prevalent in the context, for example if the word ``tiger" is prevalent in the document(s) for the question ``What is the largest living sub-species of the tiger?", greater weight will be given to question words that are less common, such as ``largest" or ``sub-species". Relative to selecting the first paragraph in the document, this improves the chance of the selected paragraph containing the correct answer from 83.1\% to 85.1\% on TriviaQA web. We also expect this approach to do a better job of selecting paragraphs that relate to the question since it is explicitly selecting paragraphs that contain question words. 

\definecolor{darkgreen}{rgb}{0, 0.65, 0}

\subsection{Handling Noisy Labels}
\label{sect:noisy-labels}
\begin{figure}[h]
\center
\fbox{\parbox{0.45\textwidth}{
\begin{small}
\textbf{Question:} Which British general was killed at Khartoum in 1885?

\textbf{Answer:} Gordon

\textbf{Context:}
In February 1885 \textcolor{red}{Gordon} returned to the Sudan to evacuate Egyptian forces. Khartoum came under siege the next month and rebels broke into the city, killing \textcolor{darkgreen}{Gordon} and the other defenders. The British public reacted to his death by acclaiming `\textcolor{red}{Gordon} of Khartoum', a saint. However, historians have suggested that \textcolor{red}{Gordon} defied orders and refused to evacuate...
\end{small}
}}
\caption{Noisy supervision causes many spans of text that contain the answer, but are not situated in a context that relates to the question, to be labelled as correct answer spans (highlighted in red). This risks distracting the model from learning from more relevant spans (highlighted in green).}
\label{fig:noisy_labels}
\end{figure}

In a distantly supervised setup we label all text spans that match the answer text as being correct. This can lead to training the model to select unwanted answer spans. Figure~\ref{fig:noisy_labels} contains an example. 
%Prior work has selected a span at random each epoch~\citep{triviaqa}, trusting that the model can overcome this problem, or heuristically selected a single answer span~\citep{hu2017mnemonic}.
To handle this difficulty, we use a summed objective function similar to the one from~\citet{kadlec2016text}, that optimizes the sum of the probabilities of all answer spans. The models we consider here work by independently predicting the start and end token of the answer span, so we take this approach for both predictions. Thus the objective for the span start boundaries becomes:
$$ -\log{\left(\frac{\sum_{k \in A} e^{s_k}}{\sum_{i=1}^n e^{s_i}}\right)} $$
where $A$ is the set of tokens that start an answer span, $n$ is the number of context tokens, and $s_i$ is a scalar score computed by the model for span $i$. This optimizes the negative log-likelihood of selecting any correct start token. 
This objective is agnostic to how the model distributes probability mass across the possible answer spans, thus the model can ``choose" to focus on only the more relevant spans. 

\subsection{Model}
\begin{figure}[!ht]
\center
\includegraphics[width=\columnwidth]{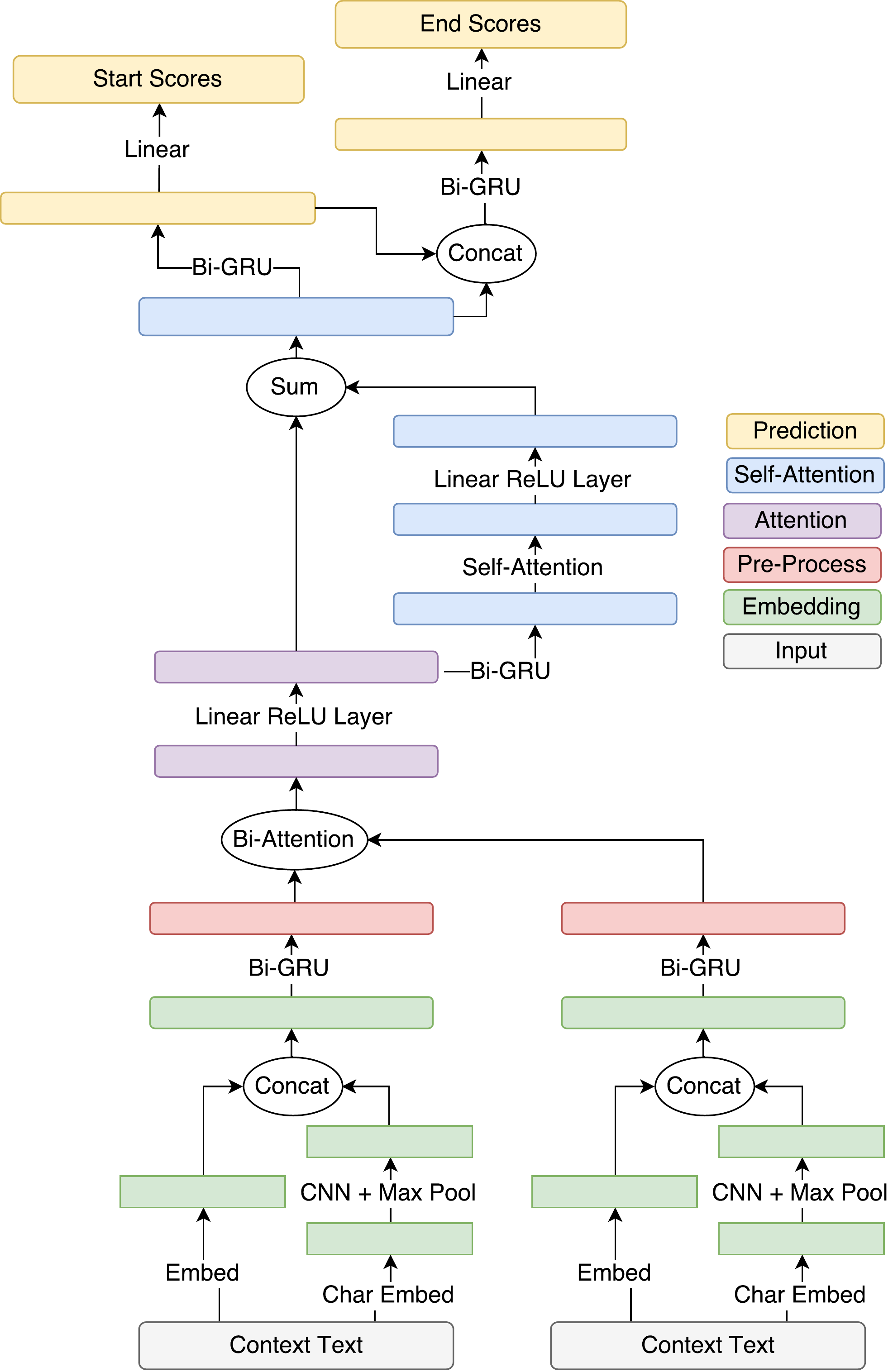}

\caption{High level outline of our model.}
\label{fig:model_outline}
\end{figure}
We use a model with the following layers (shown in Figure~\ref{fig:model_outline}):

\textbf{Embedding:}
We embed words using pre-trained word vectors. We also embed the characters in each word into size 20 vectors which are learned, and run a convolution neural network followed by max-pooling to get character-derived embeddings for each word. The character-level and word-level embeddings are then concatenated and passed to the next layer. We do not update the word embeddings during training.

\textbf{Pre-Process:}
A shared bi-directional GRU~\cite{cho2014learning} is used to map the question and passage embeddings to context-aware embeddings. 

\textbf{Attention:}
The bi-directional attention mechanism from the Bi-Directional Attention Flow (BiDAF) model~\cite{bidaf} is used to build a query-aware context representation. Let $\vec{h_i}$ be the vector for context word $i$, $\vec{q_j}$ be the vector for question word $j$, and $n_q$ and $n_c$ be the lengths of the question and context respectively. We compute attention between context word $i$ and question word $j$ as:
$$ a_{ij} = \vec{w_1}\cdot \vec{h_i} + \vec{w_2} \cdot \vec{q_j} + \vec{w_3} \cdot( \vec{h_i} \odot \vec{q_j}) $$
where $\vec{w_1}$, $\vec{w_2}$, and $\vec{w_3}$ are learned vectors and $\odot$ is element-wise multiplication. We then compute an attended vector $\vec{c_i}$ for each context token as:
$$ p_{ij} = \frac{e^{a_{ij}}}{\sum^{n_q}_{j=1}e^{a_{ij}}} $$
$$ \vec{c_i} = \sum^{n_q}_{j=1} \vec{q_j} p_{ij}$$

We also compute a query-to-context vector $\vec{q_c}$:
$$ m_i = \max_{1 \leq j \le n_q} a_{ij}$$
$$ p_i = \frac{e^{m_{i}}}{\sum^{n_c}_{i=1}e^{m_{i}}}$$
$$ \vec{q_c} = \sum^{n_c}_{i=1} \vec{h_i} p_i$$

The final vector computed for each token is built by concatenating $\vec{h_i}$, $\vec{c_i}$, $\vec{h_i} \odot \vec{c_i}$, and $\vec{q_c} \odot \vec{c_i}$. In our model we subsequently pass the result through a linear layer with ReLU activations.

\textbf{Self-Attention:}
Next we use a layer of residual self-attention. The input is passed through another bi-directional GRU. Then we apply the same attention mechanism, only now between the passage and itself. In this case we do not use query-to-context attention and we set $a_{ij} = -inf$ if $i = j$.

As before, we pass the concatenated output through a linear layer with ReLU activations. This layer is applied residually, so this output is additionally summed with the input.

\textbf{Prediction:}
In the last layer of our model a bi-directional GRU is applied, followed by a linear layer that computes answer start scores for each token. The hidden states of that layer are concatenated with the input and fed into a second bi-directional GRU and linear layer to predict answer end scores. The softmax operation is applied to the start and end scores to produce start and end probabilities, and we optimize the negative log-likelihood of selecting correct start and end tokens.

\textbf{Dropout:}
We also employ variational dropout, where a randomly selected set of hidden units are set to zero across all time steps during training~\cite{variational_dropout}. We dropout the input to all the GRUs, including the word embeddings, as well as the input to the attention mechanisms, at a rate of 0.2. 

\section{Confidence Method}
\label{sect:multi-paragraph}

\begin{table*}
\begin{small}
\begin{center}
\begin{tabular}{ | l | l | l | c | l | l | c |}
\hline 
Question & Low Confidence Correct Extraction & High Confidence Incorrect Extraction \\ 
\hline
\parbox{3cm}{When is the Members Debate held?} & \parbox{5.7cm}{\textbf{Immediately after Decision Time} a ``Members Debate" is held, which lasts for 45 minutes...} & \parbox{5.7cm}{...majority of the Scottish electorate voted for it in a referendum to be held on \textbf{1 March 1979} that represented at least...} 
\\[0.6cm]

\hline
\parbox{3cm}{How many tree species are in the rainforest?} &\parbox{5.7cm}{...plant species is the highest on Earth with one 2001 study finding a quarter square kilometer (62 acres) of Ecuadorian rainforest supports more than \textbf{1,100} tree species } & \parbox{5.7cm}{The affected region was approximately \textbf{1,160,000} square miles (3,000,000 km2) of rainforest, compared to 734,000 square miles} \\[0.6cm]

\hline
Who was Warsz? & \parbox{5.7cm}{....In actuality, Warsz was a 12th/13th century \textbf{nobleman} who owned a village located at the modern....} & \parbox{5.7cm}{One of the most famous people born in Warsaw was \textbf{Maria Sklodowska - Curie}, who achieved international...} \\[0.5cm]
\hline
\parbox{3cm}{How much did the initial LM weight in kg?} & \parbox{5.7cm}{The initial LM model weighed approximately 33,300 pounds (\textbf{15,000} kg), and...} & \parbox{5.7cm}{The module was 11.42 feet (3.48 m) tall, and 
weighed approximately 12,250 pounds (\textbf{5,560} kg)} \\[0.5cm]

\hline
\parbox{3cm}{What do the auricles do?} &\parbox{5.7cm}{...many species of lobates have four auricles, gelatinous projections edged with cilia that \textbf{produce water currents that help direct microscopic prey toward the mouth}...} & \parbox{5.7cm}{The Cestida are ribbon - shaped planktonic animals, with the mouth and aboral \textbf{organ aligned in the middle of opposite edges of the ribbon}} \\[0.6cm]	

\hline

\end{tabular}
\end{center}
\end{small}
\caption{Examples from SQuAD where a paragraph-level model was less confident in a correct extraction from one paragraph (left) than in an incorrect extraction from another (right). Even if the passage has no correct answer, the model still assigns high confidence to phrases that match the category the question is asking about. Because the confidence scores are not well-calibrated, this confidence is often higher than the confidence assigned to the correct answer span.}
\label{table:qualitative_results}
\end{table*}
 
We adapt this model to the multi-paragraph setting by using the un-normalized and un-exponentiated (i.e., before the softmax operator is applied) score given to each span as a measure of the model's confidence. For the boundary-based models we use here, a span's score is the sum of the start and end score given to its start and end token. At test time we run the model on each paragraph and select the answer span with the highest confidence. This is the approach taken by~\citet{openqa}. 

%As our experiments in Section 5 show, applying this approach with altering how the model is trained can lead to poor results. Table 1 contains some qualitative examples. 
%We hypothesize that there are two particular problem that might lead to a model's confidence scores becoming poorly calibrated. First, for models trained with the softmax objective, the pre-softmax scores for all spans can be arbitrarily increased or decreased by a constant value without changing the resulting softmax probability distribution. As a result, nothing prevents models from producing scores that are arbitrarily all larger or all smaller for one paragraph than another. Second, if the model only sees paragraphs that contain answers, it might become too confident in heuristics or patterns that are only effective when it is known a-priori that an answer exists. For example, we observe that the model can assign high confidence values to spans that strongly match the category of the answer, even if the question words do not match the context. 
%This works passably well if an answer present, but can lead to highly over-confidence extractions in other cases, as seen in Table~\ref{table:qualitative_results}. 
%Similar kinds of errors have been observed when distractor sentences are added to the context~\citet{jia2017adversarial}.
%
Applying this approach without altering how the model is trained is, however, a gamble; the training objective does not require these confidence scores to be comparable between paragraphs. Our experiments in Section~\ref{sect:experiments} show that in practice these models can be very poor at providing good confidence scores. Table~\ref{table:qualitative_results} shows some qualitative examples of this phenomenon.

We hypothesize that there are two key reasons a model's confidence scores might not be well calibrated. First, for models trained with the softmax objective, the pre-softmax scores for all spans can be arbitrarily increased or decreased by a constant value without changing the resulting softmax probability distribution. As a result, nothing prevents models from producing scores that are arbitrarily all larger or all smaller for one paragraph than another. Second, if the model only sees paragraphs that contain answers, it might become too confident in heuristics or patterns that are only effective when it is known \textit{a priori} that an answer exists. For example, in Table~\ref{table:qualitative_results} we observe that the model will assign high confidence values to spans that strongly match the category of the answer, even if the question words do not match the context. 
This might work passably well if an answer is present, but can lead to highly over-confident extractions in other cases.
Similar kinds of errors have been observed when distractor sentences are added to the context~\citep{jia2017adversarial}.
 
We experiment with four approaches to training models to produce comparable confidence scores, shown in the follow subsections. In all cases we will sample paragraphs that do not contain an answer as additional training points.

\subsection{Shared-Normalization}
In this approach all paragraphs are processed independently as usual. However, a modified objective function is used where the normalization factor in the softmax operation is shared between all paragraphs from the same context. Therefore, the probability that token $a$ from paragraph $p$ starts an answer span is computed as:
$$ \frac{e^{s_{ap}}}{\sum_{j \in P} \sum^{n_j}_{i=1} e^{s_{ij}}} $$
where $P$ is the set of paragraphs that are from the same context as $p$, and $s_{ij}$ is the score given to token $i$ from paragraph $j$. We train on this objective by including multiple paragraphs from the same context in each mini-batch.

This is similar to simply feeding the model multiple paragraphs from each context concatenated together, except that each paragraph is processed independently until the normalization step. The key idea is that this will force the model to produce scores that are comparable between paragraphs, even though it does not have access to information about the other paragraphs being considered. 

\subsection{Merge}
As an alternative to the previous method, we experiment with concatenating all paragraphs sampled from the same context together during training. A paragraph separator token with a learned embedding is added before each paragraph. Our motive is to test whether simply exposing the model to more text will teach the model to be more adept at ignoring irrelevant text.

\subsection{No-Answer Option}
We also experiment with allowing the model to select a special ``no-answer" option for each paragraph. First, note that the independent-bounds objective can be re-written as:
$$ -\log\left(\frac{e^{s_a}}{\sum^n_{i=1} e^{s_i}}\right) - \log\left(\frac{e^{g_b}}{\sum^n_{j=1} e^{g_j}}\right) = $$
%$$ -\log\left(\frac{e^{s_a g_b}}{\left(\sum^n_{i=0} e^{s_i} \right)\left(\sum^n_{j=0} e^{g_j} \right) } \right) = $$
$$ -\log\left(\frac{e^{s_a g_b}}{\sum^n_{i=1} \sum^n_{j=1} e^{s_i g_j}} \right) $$
where $s_j$ and $g_j$ are the scores for the start and end bounds produced by the model for token $j$, and $a$ and $b$ are the correct start and end tokens. We have the model compute another score, $z$, to represent the weight given to a ``no-answer" possibility. Our revised objective function becomes:
$$ 
-\log\left(\frac{(1 - \delta)e^z + \delta e^{s_a g_b}}{e^z + \sum^n_{i=1} \sum^n_{j=1} e^{s_i g_j} } \right)
$$
where $\delta$ is 1 if an answer exists and 0 otherwise. If there are multiple answer spans we use the same objective, except the numerator includes the summation over all answer start and end tokens.

We compute $z$ by adding an extra layer at the end of our model. We compute a soft attention over the span start scores, $p_i = \frac{e^{s_i}}{\sum^n_{j=1}e^ {s_j}}$, and then take the weighted sum of the hidden states from the GRU used to generate those scores, $\vec{h_i}$, giving $\vec{v_1} = \sum^n_{i=1} \vec{h_i} p_i$. We compute a second vector, $\vec{v_2}$ in the same way using the end scores. Finally, a step of learned attention is performed on the output of the Self-Attention layer that computes:
$$ a_i = \vec{w} \cdot \vec{h_i} $$
$$ p_i = \frac{e^{a_i}}{\sum^n_{j=1} e^{a_j}} $$
$$ \vec{v_3} = \sum^n_{i=1} \vec{h_i} p_i $$
where $\vec{w}$ is a learned weight vector and $\vec{h_i}$ is the vector for token $i$.

We concatenate these three vectors and use them as input to a two layer network with an 80 dimensional hidden layer and ReLU activations that produces $z$ as its only output.

\subsection{Sigmoid}
As a final baseline, we consider training models with the sigmoid loss objective function. That is, we compute a start/end probability for each token in the context by applying the sigmoid function to the start/end scores of each token. A cross entropy loss is used on each individual probability. The intuition is that, since the scores are being evaluated independently of one another, they will be comparable between different paragraphs.

\section{Experimental Setup}
\subsection{Datasets}
We evaluate our approach on three datasets: TriviaQA unfiltered~\cite{triviaqa}, a dataset of questions from trivia databases paired with documents found by completing a web search of the questions; TriviaQA web, a dataset derived from TriviaQA unfiltered by treating each question-document pair where the document contains the question answer as an individual training point; and SQuAD~\cite{squad}, a collection of Wikipedia articles and crowdsourced questions. 

\subsection{Preprocessing}
We note that for TriviaQA web we do not subsample as was done by~\citet{triviaqa}, instead training on the full 530k question-document training pairs. We also observed that the metrics for TriviaQA are computed after applying a small amount of text normalization (stripping punctuation, removing articles, ect.) to both the ground truth text and the predicted text. As a result, some spans of text that would have been considered an exact match after normalization were not marked as answer spans during preprocessing, which only detected exact string matches. We fix this issue by labeling all spans of text that would have been considered an exact match by the official evaluation script as an answer span.

In TriviaQA, documents often contain many small paragraphs, so we merge paragraphs together as needed to get paragraphs of up to a target size. We use a maximum size of 400 unless stated otherwise. Paragraph separator tokens with learned embeddings are added between merged paragraphs to preserve formatting information.

\subsection{Sampling}
Our confidence-based approaches are all trained by sampling paragraphs, including paragraphs that do not contain an answer, during training. For SQuAD and TriviaQA web we take the top four paragraphs ranked by TF-IDF score for each question-document pair. We then sample two different paragraphs from this set each epoch. Since we observe that the higher-ranked paragraphs are much more likely to contain the context needed to answer the question, we sample the highest ranked paragraph that contains an answer twice as often as the others. For the merge and shared-norm approaches, we additionally require that at least one of the paragraphs contains an answer span. 

For TriviaQA unfiltered, where we have multiple documents for each question, we find it beneficial to use a more sophisticated paragraph ranking function. In particular, we use a linear function with five features: the TF-IDF cosine distance, whether the paragraph was the first in its document, how many tokens occur before it, and the number of case insensitive and case sensitive matches with question words. The function is trained on the distantly supervised objective of selecting paragraphs that contain at least one answer span. We select the top 16 paragraphs for each question and sample pairs of paragraphs as before.

\subsection{Implementation}
We train the model with the Adadelta optimizer~\cite{zeiler2012adadelta} with a batch size 60 for TriviaQA and 45 for SQuAD. At test time we select the most probable answer span of length less than or equal to 8 for TriviaQA and 17 for SQuAD. The GloVe 300 dimensional word vectors released by~\citet{pennington2014glove} are used for word embeddings. On SQuAD, we use a dimensionality of size 100 for the GRUs and of size 200 for the linear layers employed after each attention mechanism. We find for TriviaQA, likely because there is more data, using a larger dimensionality of 140 for each GRU and 280 for the linear layers is beneficial. During training, we maintain an exponential moving average of the weights with a decay rate of 0.999. We use the weight averages at test time.

\section{Results}
\label{sect:experiments}

\subsection{TriviaQA Web}
\setlength\doublerulesep{.8pt}

\begin{table}

\center
\begin{tabular}{l c c}
\hhline{===}
Model & EM & F1 \\ \hline
baseline~\cite{triviaqa} & 41.08 & 47.40 \\ 
BiDAF & 50.21 & 56.86\\ 
BiDAF + TF-IDF & 53.41 & 59.18\\ 
BiDAF + sum & 56.22 & 61.48\\ 
BiDAF + TF-IDF + sum & 57.20 & 62.44\\ 
our model + TF-IDF + sum & 61.10 & 66.04\\ 
\hhline{===}
\end{tabular}
\caption{Results on TriviaQA web using our pipelined method. We significantly improve upon the baseline by combining the preprocessing procedures, TF-IDF paragraph selection, the sum objective, and our model design.}
\label{table:bidaf-ablate}
\end{table}

\begin{figure*}
\center

\begin{minipage}[b]{.49\textwidth}
\includegraphics[width=\columnwidth]{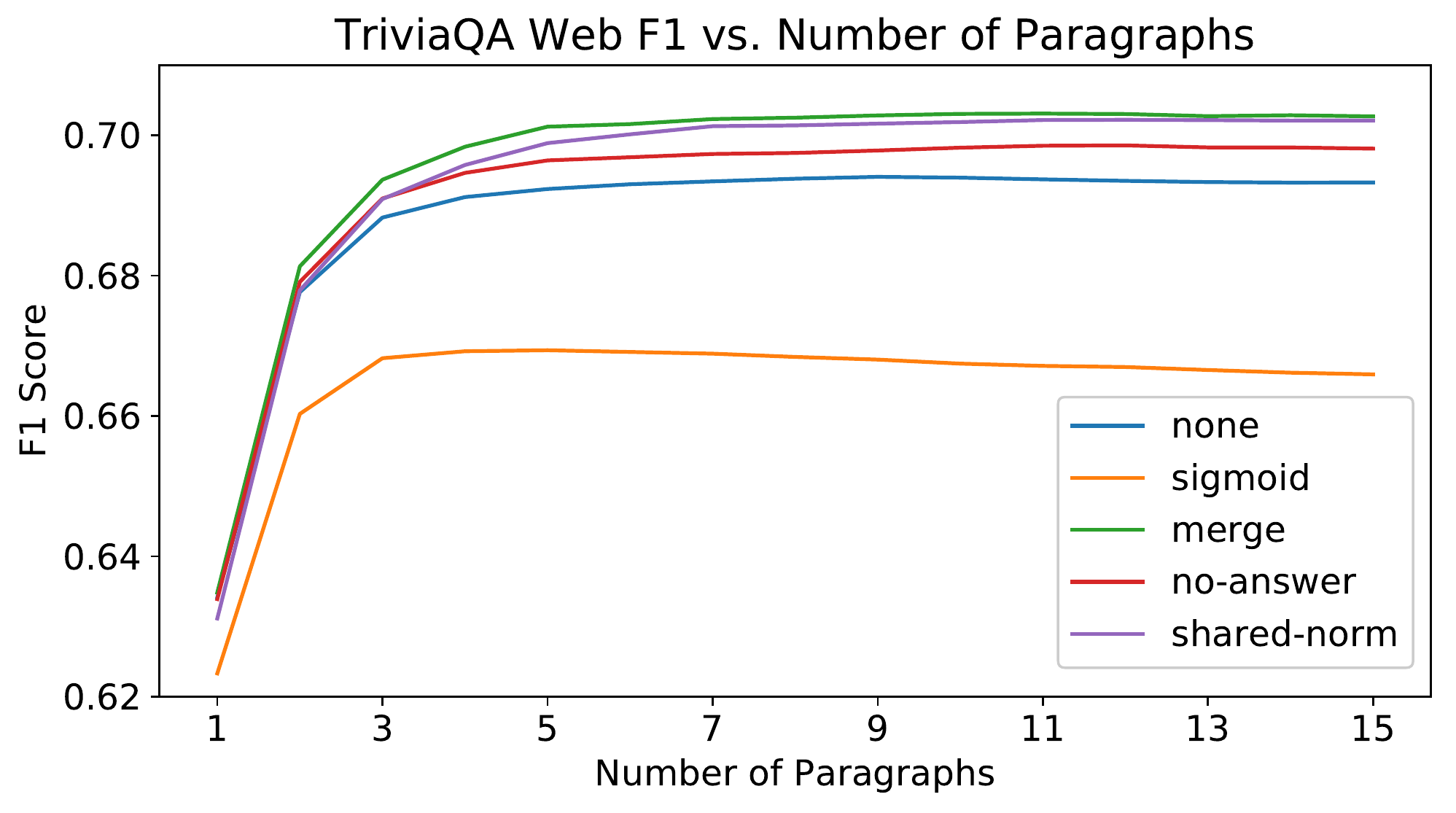}
\end{minipage}
\hfill
\begin{minipage}[b]{.49\textwidth}
\includegraphics[width=\columnwidth]{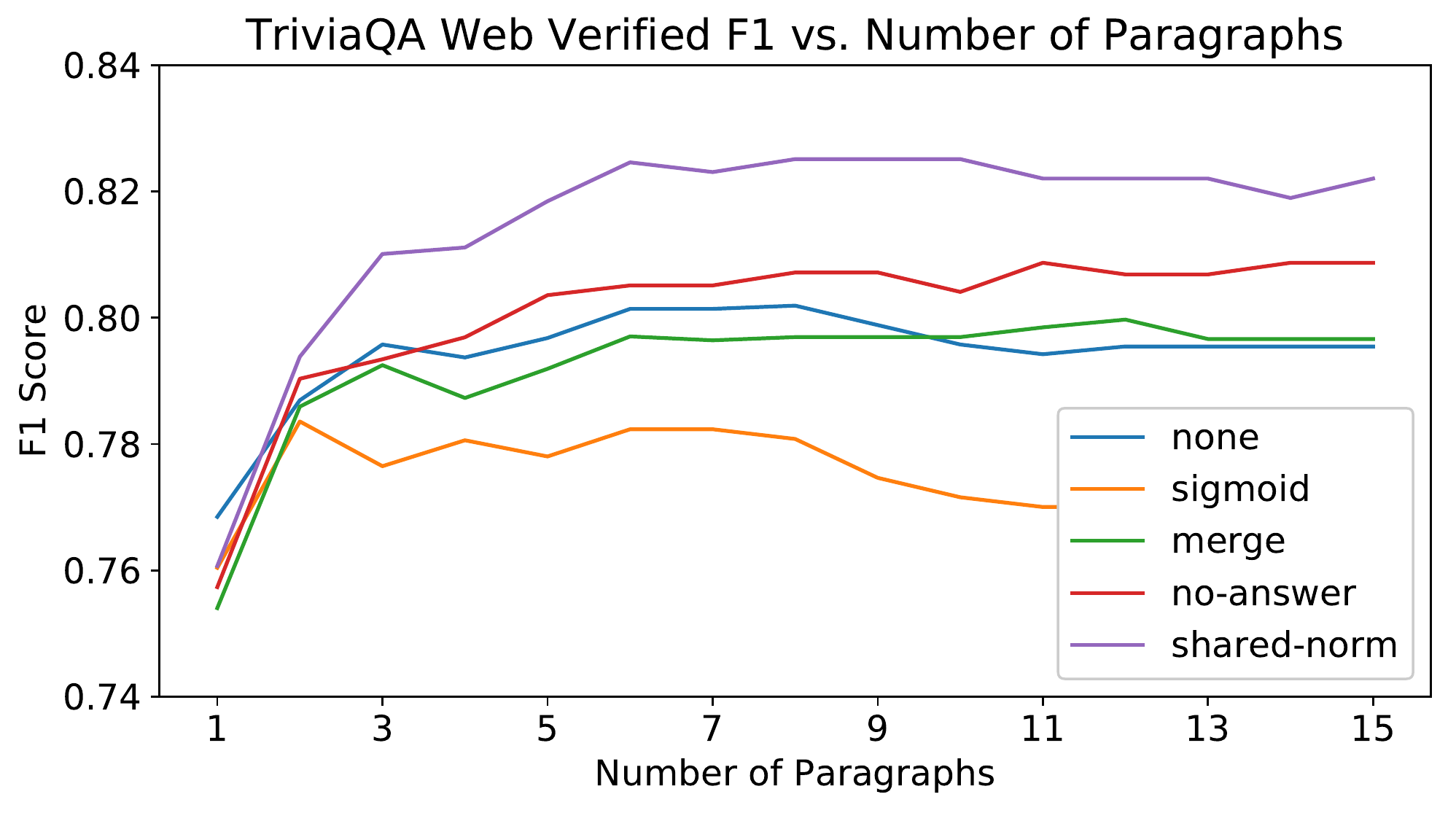}
\end{minipage}

\caption{Results on TriviaQA web (left) and verified TriviaQA web (right) when applying our models to multiple paragraphs from each document. The shared-norm, merge, and no-answer training methods improve the model's ability to utilize more text, with the shared-norm method being significantly ahead of the others on the verified set and tied with the merge approach on the general set.}
\label{fig:web-document-results}
\end{figure*}

\begin{table*}
\setlength\doublerulesep{.8pt}
\center
\begin{tabular}{l | c | c | c | c }
\hhline{=====}
Model & \multicolumn{2}{c}{All} & \multicolumn{2}{c}{Verified} \\ 
 & EM & F1 & EM & F1 \\ \hline
 baseline~\cite{triviaqa} & 40.74 & 47.06 & 49.54 & 55.80 \\
MEMEN*~\cite{pan2017memen} & 43.16 & 46.90 & 49.28 & 55.83 \\
Mnemonic Reader~\cite{hu2017mnemonic} & 46.94 & 52.85 & 54.45 & 59.46 \\
Reading Twice for NLU~\cite{weissenborn2017reading} & 50.56 & 56.73 & 63.20 & 67.97 \\ \hline
S-Norm (ours) & 66.37 & 71.32 & 79.97 & 83.70 \\
\hhline{=====}
 \multicolumn{5}{l}{\footnotesize{*Results on the dev set}} \\

\end{tabular}
\caption{Published TriviaQA results. We advance the state of the art by about 15 points both test sets. }
\label{table:triviaqa-leader-board}
\end{table*}

First, we do an ablation study on TriviaQA web to show the effects of our proposed methods for our pipeline model. We start with an implementation of the baseline from~\cite{triviaqa}. Their system selects paragraphs by taking the first 400 tokens of each document, uses BiDAF~\cite{bidaf} as the paragraph model, and selects a random answer span from each paragraph each epoch to be used in BiDAF's cross entropy loss function during training. Paragraphs of size 800 are used at test time. As shown in Table~\ref{table:bidaf-ablate}, our implementation of this approach outperforms the results reported by~\citet{triviaqa} significantly, likely because we are not subsampling the data. We find both TF-IDF ranking and the sum objective to be effective; even without changing the model we achieve state-of-the-art results. Using our refined model increases the gain by another 4 points.

Next we show the results of our confidence-based approaches. In this setting we group each document's text into paragraphs of at most 400 tokens and rank them using our TF-IDF heuristic. Then we measure the performance of our proposed approaches as the model is used to independently process an increasing number of these paragraphs and the model's most confident answer is returned. We additionally measure performance on the verified portion of TriviaQA, a small subset of the question-document pairs in TriviaQA web where humans have manually verified that the document contains sufficient context to answer the question. The results are shown in Figure~\ref{fig:web-document-results}. 

On these datasets even the model trained without any of the proposed training methods (``none") improves as it is allowed to use more text, showing it does a passable job at focusing on the correct paragraph. The no-answer option training approach lead to a significant improvement, and the shared-norm and merge approach are even better. On the verified set, the shared-norm approach is solidly ahead of the other options. This suggests the shared-norm model is better at extracting answers when it is clearly stated in the text, but worse at guessing the answer in other cases. 

We use the shared-norm approach for evaluation on the TriviaQA test set. We found that increasing the paragraph size to 800 at test time, and re-training the model on paragraphs of size 600, was slightly beneficial, allowing our model to reach 66.04 EM and 70.98 F1 on the dev set. We submitted this model to be evaluated on the TriviaQA test set and achieved 66.37 EM and 71.32 F1, firmly ahead of prior work, as shown in Table~\ref{table:triviaqa-leader-board}. Note that human annotators have estimated that only 75.4\% of the question-document pairs contain sufficient evidence to answer the question~\cite{triviaqa}, which suggests we are approaching the upper bound for this task. However, the score of 83.7 F1 on the verified set suggests that there is still room for improvement.

\subsection{TriviaQA Unfiltered}

\begin{figure}[!ht]
\includegraphics[width=\columnwidth]{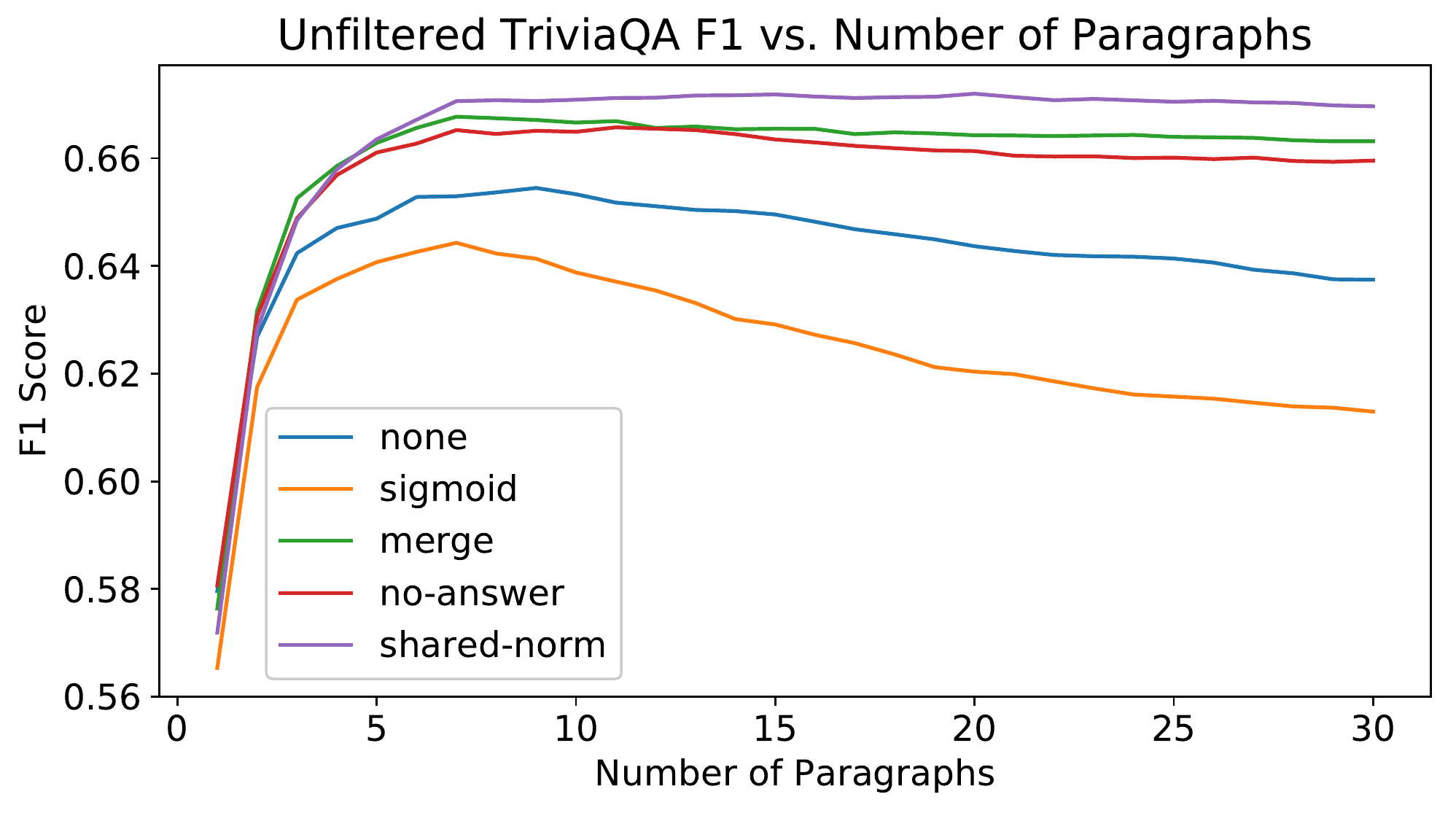}
\caption{Results for our confidence methods on TriviaQA unfiltered. Here we see a more dramatic difference between these models. The shared-norm approach is the strongest, while the base model starts to lose performance as more paragraphs are used.}
\label{fig:open-results}
\end{figure}

Next we apply our confidence methods to TriviaQA unfiltered. 
This dataset is of particular interest because the system is not told which document contains the answer, so it provides a plausible simulation of attempting to answer a question using a document retrieval system.
We show the same graph as before for this dataset in Figure~\ref{fig:open-results}. On this dataset it is more important to train the model to produce well calibrated confidence scores. Note the base model starts to lose performance as more paragraphs are used, showing that errors are being caused by the model being overly confident in incorrect extractions. 

\subsection{SQuAD}
\addtocounter{footnote}{-1}

\label{sect:squad_results}

\label{sect:squad-results}
\begin{table}
\center
\begin{tabular}{ c | c | c | c | c}
\hhline{=====}
& \multicolumn{2}{c}{Dev} & \multicolumn{2}{c}{Test} \\
Model & EM & F1 & EM & F1 \\ \hline
none & 71.60 & 80.78 & 72.14 & 81.05 \\
sigmoid & 70.28 & 79.05 & - & -\\
merge & 71.20 & 80.26 & - & -\\
no-answer & 71.51 & 80.71 & - & -\\
shared-norm & 71.16 & 80.23 & - & -\\
\hhline{=====}
\end{tabular}
\caption{
Results on the standard SQuAD dataset. The test scores place our model as 8th on the SQuAD leader board among non-ensemble models\footnotemark. Training with the proposed multi-paragraph approaches only leads to a marginal drop in performance in this setting.}
\label{table:squad-paragraph}
\end{table}

\begin{figure}
\includegraphics[width=\columnwidth]{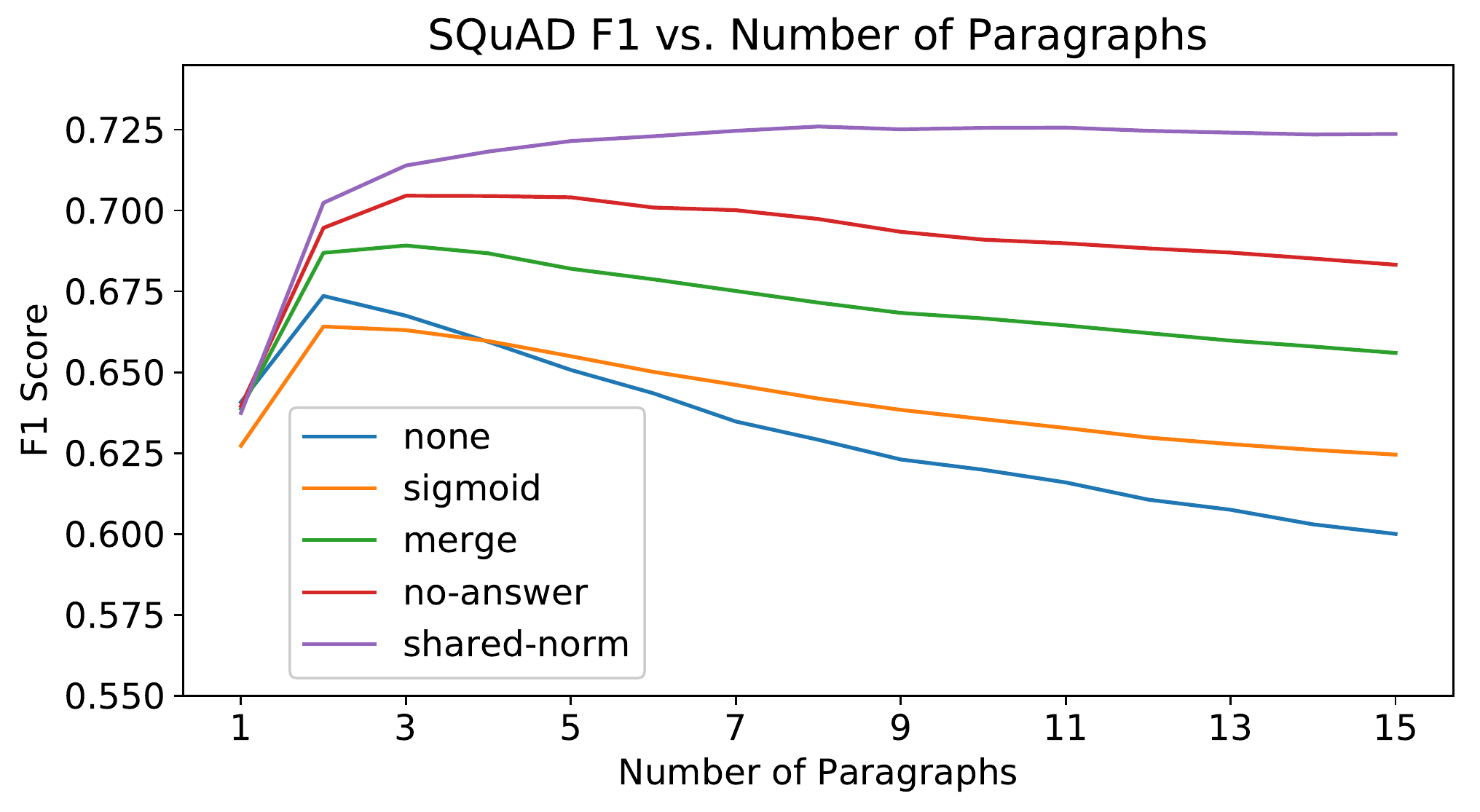}
\caption{Results for our confidence methods on document-level SQuAD. The base model does poorly in this case, rapidly losing performance once more than two paragraphs are used. While all our approaches had some benefit, the shared-norm model is the strongest, and is the only one to not lose performance as large numbers of paragraphs are used.}
\label{figure:squad_plot}
\end{figure}

We additionally evaluate our model on SQuAD. SQuAD questions were not built to be answered independently of their context paragraph, which makes it unclear how effective of an evaluation tool they can be for document-level question answering. To assess this we manually label 500 random questions from the training set. 
%
%We categorize questions as \textbf{(i)}: Context-independent, meaning it can be understood independently of the paragraph. \textbf{(ii)}: Document-dependent, meaning it can be understood given the article's title. For example, ``What individual is the school named after?" for the document ``Harvard University". \textbf{(iii)}: Paragraph-dependent, meaning it can only be understood in the context of the paragraph. For example, ``What was the first step in the reforms?".
%
We categorize questions as:
\begin{enumerate}
\item Context-independent, meaning it can be understood independently of the paragraph.
\item Document-dependent, meaning it can be understood given the article's title. For example, ``What individual is the school named after?" for the document ``Harvard University".
\item Paragraph-dependent, meaning it can only be understood given its paragraph. For example, ``What was the first step in the reforms?".
\end{enumerate}

\footnotetext{as of 10/23/2017}

We find 67.4\% of the questions to be context-independent, 22.6\% to be document-dependent, and the remaining 10\% to be paragraph-dependent. The many document-dependent questions stem from the fact that questions are frequently about the subject of the document, so the article's title is often sufficient to resolve co-references or ambiguities that appear in the question. Since a reasonably high fraction of the questions can be understood given the document they are from, and to isolate our analysis from the retrieval mechanism used, we choose to evaluate on the document-level. We build documents by concatenating all the paragraphs in SQuAD from the same article together into a single document. 

The performance of our models given the correct paragraph (i.e., in the standard SQuAD setting), is shown in Table~\ref{table:squad-paragraph}. Our paragraph-level model is competitive on this task, and our variations to handle the multi-paragraph setting only cause a minor loss of performance. 

We graph the document-level performance in Figure~\ref{figure:squad_plot}. For SQuAD, we find it crucial to employ one of the suggested confidence training techniques. The base model starts to drop in performance once more than two paragraphs are used. However, the shared-norm approach is able to reach a peak performance of 72.37 F1 and 64.08 EM given 15 paragraphs. Given our estimate that 10\% of the questions are ambiguous if the paragraph is unknown, our approach appears to have adapted to the document-level task very well.

Finally, we compare the shared-norm model with the document-level result reported by~\citet{openqa}. We re-evaluate our model using the documents used by~\citet{openqa}, which consist of the same Wikipedia articles SQuAD was built from, but downloaded at different dates. The advantage of this dataset is that it does not allow the model to know \textit{a priori} which paragraphs were filtered out during the construction of SQuAD. The disadvantage is that some of the articles have been edited since the questions were written, so some questions may no longer be answerable. Our model achieves 59.14 EM and 67.34 F1 on this dataset, which significantly outperforms the 49.7 EM reported by~\citet{openqa}. 

\subsection{Discussion}
We found that models that have only been trained on answer-containing paragraphs can perform very poorly in the multi-paragraph setting. The results were particularly bad for SQuAD, we think this is partly because the paragraphs are shorter, so the model had less exposure to irrelevant text. In general, we found the shared-norm approach to be the most effective way to resolve this problem. 
%We suspect that the advantage it has over the merge approach is that the merge-trained model is used to ``seeing"
%We suspect that the key advantage it has is that its trains the model to avoid ``guessing" at an answer when there is little evidence because this would risk overriding its answer from a richer paragraph. 
The no-answer and merge approaches were moderately effective, but we note that they do not resolve the scaling problem inherent to the softmax objective we discussed in Section~\ref{sect:multi-paragraph}, which might be why they lagged behind. 
The sigmoid objective function reduces the paragraph-level performance considerably, especially on the TriviaQA datasets. We suspect this is because it is vulnerable to label noise, as discussed in Section~\ref{sect:noisy-labels}.

\section{Related Work}
\textbf{Reading Comprehension Datasets.} The state of the art in reading comprehension has been rapidly advanced by neural models, in no small part due to the introduction of many large datasets. The first large scale datasets for training neural reading comprehension models used a Cloze-style task, where systems must predict a held out word from a piece of text~\cite{hermann2015teaching, hill2015goldilocks}. Additional datasets including SQuAD~\cite{squad}, WikiReading~\cite{hewlett2016wikireading}, MS Marco~\cite{nguyen2016ms} and TriviaQA~\cite{triviaqa} provided more realistic questions. Another dataset of trivia questions, Quasar-T~\cite{dhingra2017quasar}, was introduced recently that uses ClueWeb09~\cite{callan2009clueweb09} as its source for documents. In this work we choose to focus on SQuAD and TriviaQA.

\textbf{Neural Reading Comprehension.} 
Neural reading comprehension systems typically use some form of attention~\citep{wang2016machine}, although alternative architectures exist~\cite{openqa, weissenborn2017fastqa}. Our model follows this approach, but includes some recent advances such as variational dropout~\cite{variational_dropout} and bi-directional attention~\cite{bidaf}. Self-attention has been used in several prior works~\cite{cheng2016long, wang2017gated, pan2017memen}. Our approach to allowing a reading comprehension model to produce a per-paragraph no-answer score is related to the approach used in the BiDAF-T~\cite{min2017question} model to produce per-sentence classification scores, although we use an attention-based method instead of max-pooling.

\textbf{Open QA.}
Open question answering has been the subject of much research, especially spurred by the TREC question answering track~\cite{voorhees1999trec}. Knowledge bases can be used, such as in~\cite{berant2013semantic}, although the resulting systems are limited by the quality of the knowledge base. Systems that try to answer questions using natural language resources such as YodaQA~\cite{baudivs2015yodaqa} typically use pipelined methods to retrieve related text, build answer candidates, and pick a final output.

\textbf{Neural Open QA.}
Open question answering with neural models was considered by~\citet{openqa}, where researchers trained a model on SQuAD and combined it with a retrieval engine for Wikipedia articles. Our work differs because we focus on explicitly addressing the problem of applying the model to multiple paragraphs. A pipelined approach to QA was recently proposed by~\citet{wang2017r}, where a ranker model is used to select a paragraph for the reading comprehension model to process. 

\section{Conclusion}
We have shown that, when using a paragraph-level QA model across multiple paragraphs, our training method of sampling non-answer containing paragraphs while using a shared-norm objective function can be very beneficial.
Combining this with our suggestions for paragraph selection, using the summed training objective, and our model design allows us to advance the state of the art on TriviaQA by a large stride. 
As shown by our demo, this work can be directly applied to building deep learning powered open question answering systems.

\bibliography{paper}
\bibliographystyle{acl_natbib.bst}

\end{document}